\documentclass[letterpaper, 10 pt, conference]{ieeeconf}  % Comment this line out if you need a4paper
\let\labelindent\relax

\IEEEoverridecommandlockouts                              % This command is only needed if 
\usepackage[hyphens]{url}

\usepackage{balance}
\usepackage{enumitem}

\usepackage{booktabs}
\usepackage{multirow}
\usepackage[normalem]{ulem}
\useunder{\uline}{\ul}{}
\usepackage{threeparttable}

\usepackage[ruled,linesnumbered]{algorithm2e}

\usepackage{cite}
\usepackage{amsmath,amssymb,amsfonts}
\usepackage{algorithmic}
\usepackage{graphicx}
\usepackage{textcomp}
\usepackage{xcolor}

\usepackage{subcaption}

\makeatletter
\let\NAT@parse\undefined
\makeatother
\usepackage{hyperref}

\title{\LARGE \bf
Efficient Multi-Task Modeling \\through Automated Fusion of Trained Models
}

\author{Jingxuan Zhou, Weidong Bao*, Ji Wang, Zhengyi Zhong and Dayu Zhang% <-this % stops a space
\thanks{*This work was supported in part by the National Natural Science Foundation of China under Grants 62002369 and 62102445, in part by the Postgraduate Scientific Research Innovation Project of Hunan Province Grant XJCX2023013.}% <-this % stops a space
\thanks{Jingxuan Zhou, Weidong Bao, Ji Wang, Zhengyi Zhong and Dayu Zhang are with Laboratory for Big Data and Decision, National University of Defense Technology ChangSha 410073, China (e-mail: \{zhoujingxuan, wdbao\}@nudt.edu.cn).}
}

\begin{document}

\maketitle
\thispagestyle{empty}
\pagestyle{empty}

%%%%%%%%%%%%%%%%%%%%%%%%%%%%%%%%%%%%%%%%%%%%%%%%%%%%%%%%%%%%%%%%%%%%%%%%%%%%%%%%
\begin{abstract}

Although multi-task learning is widely applied in intelligent services, traditional multi-task modeling methods often require customized designs based on specific task combinations, resulting in a cumbersome modeling process. Inspired by the rapid development and excellent performance of single-task models, this paper proposes an efficient multi-task modeling method that can automatically fuse trained single-task models with different structures and tasks to form a multi-task model.
As a general framework, this method allows modelers to simply prepare trained models for the required tasks, simplifying the modeling process while fully utilizing the knowledge contained in the trained models. This eliminates the need for excessive focus on task relationships and model structure design. To achieve this goal, we consider the structural differences among various trained models and employ model decomposition techniques to hierarchically decompose them into multiple operable model components. Furthermore, we have designed an Adaptive Knowledge Fusion (AKF) module based on Transformer, which adaptively integrates intra-task and inter-task knowledge based on model components.
Through the proposed method, we achieve efficient and automated construction of multi-task models, and its effectiveness is verified through extensive experiments on three datasets.
Our code and related baseline methods can be found at: \href{https://github.com/zxccvdql/EMM}{https://github.com/zxccvdql/EMM}.

\end{abstract}

%%%%%%%%%%%%%%%%%%%%%%%%%%%%%%%%%%%%%%%%%%%%%%%%%%%%%%%%%%%%%%%%%%%%%%%%%%%%%%%%
\section{INTRODUCTION}
As artificial intelligence increasingly becomes the mainstream of intelligent services, the significance of Multi-Task Learning (MTL) methods has become increasingly prominent. These methods can handle multidimensional information and meet diverse user needs while saving computational resources, thus gradually becoming a key technology for intelligent models~\cite{Ma2018,Zhao2019}.
As an important branch of machine learning, MTL allows a single model to simultaneously process and output multiple tasks. By utilizing domain-specific information from related task training signals, MTL can effectively enhance the model's generalization ability~\cite{Caruana1997}. This advantage significantly improves the model's performance in intelligent services.
However, current MTL modeling methods often require the design of differentiated model structures for different task relationships (such as parallel, cascade, or auxiliary tasks)~\cite{Wang2023}. In intelligent services, user task requirements are often diverse. This leads to the traditional MTL modeling process consuming a significant amount of human resources, severely affecting the efficiency of MTL model construction, and thus restricting the ability of artificial intelligence to provide a wide range of multi-tasking services.

Inspired by the rapid advancements and outstanding performance of single-task models, we pose an intriguing question: \textit{Can we quickly construct a multi-task model by integrating these trained single-task models?} To address this question, we developed a flexible \textbf{E}fficient \textbf{M}ulti-Task \textbf{M}odeling (EMM) framework, which aims to achieve efficient automation of the multi-task modeling process through the fusion of trained single-task models.
To maintain brevity and consistency in terminology, we use the term "trained models" specifically to refer to "trained single-task models." During the design of the EMM framework, we encountered the following challenges:

\begin{figure*}[htbp]
\includegraphics[width=\textwidth]{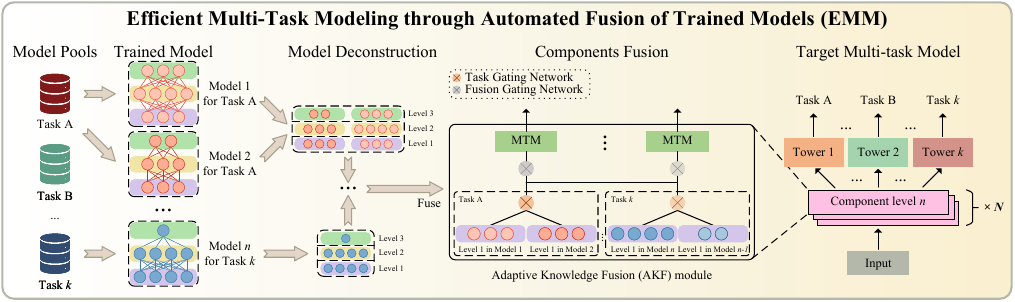}
\caption{Illustrates Efficient Multi-Task Modeling (EMM) framework. When presenting the model pool and model structures, we employ different colors to distinguish various tasks and utilize diverse model styles to represent various model structures. Specifically, model structures with the same background color indicate that this part of the model shares the same input and output characteristics. The core advantage of the EMM method lies in its ability to deconstruct trained models into multiple independent model components and combine them hierarchically using the AKF module. For the same task, this method employs the MoE ideology to achieve knowledge fusion; whereas for different tasks, it adopts the Multi-Task Merge (MTM) method for fusion after adaptive screening. Subsequently, by stacking these component layers, we can effectively achieve multi-task modeling.}
\label{fig1}
\end{figure*}

\begin{itemize}[leftmargin=*]
\item We have designed and proposed a multi-task modeling framework, named EMM, which aims to achieve efficiency in multi-task modeling. By organically combining multiple trained models, EMM can rapidly construct a high-performance multi-task model without considering the interrelationships between tasks. In this process, the selection of training models is not constrained by task types or structures, providing great flexibility.
\item To address the heterogeneity among different trained models, we have devised a decomposition method for trained models. This approach enables the hierarchical decomposition of each trained model into multiple easily integrable components, thereby facilitating knowledge fusion among heterogeneous models both within and across tasks.
\item We design an AKF module based on the self-attention mechanism to facilitate knowledge sharing and fusion within and between tasks. Specifically, for intra-task knowledge fusion, we adopt the MoE method. For inter-task knowledge fusion, we use an adaptive approach to identify the two most relevant tasks and employ the self-attention mechanism to achieve effective knowledge fusion.
% \item To validate the effectiveness of the EMM framework, we conduct experiments on three multi-task datasets. The experimental results fully demonstrate the superiority of the EMM framework in various scenarios. Furthermore, ablation studies and other experiments further verify the rationality and effectiveness of our design methods.
\end{itemize}

\section{Related Work}
\subsection{Multi-task Learning}

MTL aims to learn multiple tasks simultaneously to maximize overall performance~\cite{Zhang2022}. Research is divided into hard and soft parameter sharing~\cite{Vandenhende2022}.
Hard parameter sharing, early MTL's core approach~\cite{Jou2016,Huang2015}, shares parameters across tasks to reduce resource use but is prone to negative transfer~\cite{Kang2011}.
Soft parameter~\cite{Ma2018,Qin2020,Tang2020,Xi2021,Li2023} sharing assigns unique parameters to each task, reducing negative transfer. 

Our EMM method aligns with soft parameter sharing, enabling tasks to retain efficacy while drawing knowledge from others, ensuring exceptional individual performance.

\subsection{Model Fusion}
Model fusion is a well-studied field, mainly divided into traditional model integration (combining predictions from multiple models to enhance performance~\cite{Jiang2023}) and weight merging (fusing models at the parameter level, e.g., Gupta et al.~\cite{Gupta2020} and Wortsman et al.~\cite{Wortsman2022} merging similar models).
Researchers like Rame et al.~\cite{Rame2022}, and Arpit et al.~\cite{Arpit2022} explored merging models trained under different settings, analyzing data distribution and proposing strategies for better generalization. Jin et al.~\cite{Jin2023} aimed to unify models trained on different datasets into a robust model. Others, such as Huang et al.~\cite{Huang2024}, and Zhang et al.~\cite{Zhang2023}, used linear operations to optimize adapter parameters, enhancing generalization.

While current fusion methods integrate multiple models effectively, weight merging is usually limited to identical-structured models for a single task. Our approach, however, offers flexibility, combining knowledge from diverse models regardless of structure or task, to create a versatile model capable of multiple tasks.

\section{Methodology}
The specific steps of our method are as follows:
Collect Trained Models: First, for each task, we collect a sufficient number of trained models from various model repositories. These models may have different structures.
Deconstruct Trained Models: Next, we deconstruct each collected trained model into a set of model components to facilitate more flexible knowledge integration in subsequent steps.
Hierarchically Integrate Model Components: Using the AKF module, we hierarchically integrate the deconstructed model components. This step aims to achieve knowledge sharing both within the same task and across different tasks.
Construct Multi-Task Model: Finally, we stack multiple AKF modules to construct the required multi-task model according to specific needs.
The deconstruction process of trained models is detailed in Section~\ref{sec3}; an in-depth explanation of the AKF module can be found in Section~\ref{sec4}; and a detailed description of the EMM modeling process is provided in Section~\ref{sec5}.

\subsection{Deconstruction of Trained Models}\label{sec3}
Within the EMM framework, the cornerstone of multi-task modeling lies in the effective utilization of trained models.
However, the diversity of task types and differences in model structures pose challenges for the direct application of these models. To address this challenge, we propose a novel approach for decomposing trained models.
The core idea of this method is to split each trained model into several hierarchical model components, ensuring that components from different models have consistent input-output characteristics at the same level.

Specifically, the set of trained models can be represented as $M = \left\lbrack {m_{1}, m_{2}, \ldots, {m}_{N}} \right\rbrack$, where $N = |M|$ represents the number of models, and $m_{n}$ represents the $n$-th trained model ($n \in N$).
Correspondingly, the set of layer structures of these models can be represented as $L = \left\lbrack {l_{1}, l_{2}, \ldots, {l}_{N}} \right\rbrack$, where $l_{n}$ represents the layer structure of model $m_{n}$.
Our goal is to identify layers with identical structures across all models in set $L$ and deconstruct the models based on these identical layers, breaking down all trained models into components with the same number of hierarchical levels.
For example, consider model $A$ ($18$ layers) and model $B$ ($24$ layers). If there is only one layer with an identical structure between them, such as the $8$th layer in model $A$ and the $20$th layer in model $B$, we will mark these layers and deconstruct each model into two components accordingly. The detailed deconstruction steps are provided in Algorithm~\ref{alg1}.
\begin{algorithm}[htbp]
\SetAlgoLined %显示end
\caption{Deconstruction of Trained Models}%算法名字
\label{alg1}
\KwIn{Initial same structural layer set: $s \gets [l_1]$,
Trained model collection: $L \gets \left\lbrack {l_{1},l_{2},\ldots,{l}_{N}} \right\rbrack$,
Trained model structure set: $ M \gets \left\lbrack {m_{1},m_{2},\ldots{,m}_{N}} \right\rbrack $,
Model component parameters: $T \gets \emptyset $,
Initialize model component set: $M_{C} \gets \emptyset $.}%输入参数
\KwOut{All model components of set $M$: $M_{C}$.}%输出
% some description\; %\;用于换行
Find the same structure layer set. \\
\For{$l = l_{1},l_{2},l_{3},\ldots$}{
$\left. c\leftarrow s \cap l \right.$ \tcp*{Find the same structure}
\If{$l = \emptyset$}{\textbf{break}}
}
Decompose the model to get components.\\
\For{$m = m_{1},m_{2},m_{3},\ldots$}{
  \For{$o = o_{1},o_{2},o_{3},\cdots$ \tcp*{$o \in l$, single layer structure in model $m$}}{
$T \Leftarrow o$ \tcp*{Append element $o$ to list $T$}
\If{$o \in c$}{
  $m_{c} \gets T$ \tcp*{Append layer}
  $T \gets \emptyset $}
}
$M_{C} \Leftarrow m_{c}$ \tcp*{Append model components}
$m_{c} \gets \emptyset$
}
\end{algorithm}

The core of the deconstruction algorithm consists of two main steps. First, the algorithm performs a layer-by-layer scan of all collected trained models to identify a set of common structural layers, denoted as $c$. This set is obtained by computing the intersection of the layer structures of all models, mathematically expressed as $ (l_1 \cap l_2 \cap \ldots \cap l_N) $. The scan starts from the layer structure of any model and progressively filters out the common structural layers present in all trained models.
Second, based on the common structural layers in set $c$, the algorithm segments each trained model. During this process, the algorithm re-examines the layer structures of each model and uses the layers in set $c$ as demarcation points to divide the models into several components.
The set of components derived from the same model is uniformly represented as $m_{c}$, and these components are then aggregated to form a complete set of model components, denoted as $M_{C}$.
It is worth noting that the number of components decomposed from each trained model is equal, meaning each model will produce $|m_{c}|$ components, and these components are divided into different hierarchical levels according to the order of segmentation. Components at the same hierarchical level, although originating from different trained models, share the same input and output characteristics, facilitating subsequent model fusion operations.

This method is based on a core assumption that there are identical layer structures in all selected trained models. This assumption indeed imposes certain limitations on the implementation of the method. However, given the current trend of standardized model construction processes and the combinatorial use of specific model architectures in the modeling process~\cite{Zhang2023a}, the likelihood of models not containing identical layer structures has been greatly reduced. Therefore, the proposed model deconstruction method still demonstrates broad application prospects.

\subsection{Adaptive Knowledge Fusion Module} \label{sec4}
\begin{figure}[htbp]
\centering
\includegraphics[width=6cm]{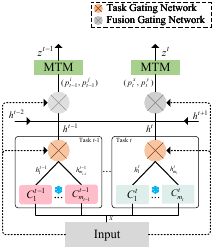}
\caption{The process of knowledge fusion is described. The task gating network is responsible for merging the knowledge from model components of the same task and the same hierarchical level. The fusion gating network selects model components from different tasks but at the same hierarchical level. After the selection is completed, the Multi-Task Merge (MTM) method is used to integrate the knowledge from components originating from different tasks.} \label{fig2}
\end{figure}

The AKF module forms the core component of the EMM framework, enabling cross-task knowledge fusion and transfer through the integration of model components in $M_{C}$. Detailed structural insights are presented in Figure~\ref{fig2}. 

As discussed in~\ref{sec3}, Algorithm~\ref{alg1} achieves the decomposition of each trained model into model components with an equal number.
These model components at the same hierarchical level follow a unified input-output specification. We integrate these components through the following steps: Given an input vector $x$, we embed it into the first-level set of model components, denoted as $C_{m_{T}}^{T} = [{C_{m_{1}}^{1}, C_{m_{2}}^{2}, \ldots, C_{m_{|T|}}^{|T|}}]$. Here, $T$ represents the task set, and $m_t$ refers to the total number of trained models in task $t$ (satisfying $\sum_{t \in T}{\left| m_{t} \right| = \left| m_{T} \right|}$).
In the subsequent training process, the parameters of these model components remain frozen.
For task $t$ (where $t \in T$), its set of model components $C_{m_{t}}^{t}$ includes all model components generated by the trained models performing task $t$. Therefore, the output of the first-level model components can be expressed as:
\begin{equation}
  h = \left\lbrack {h_{1}^{1},h_{2}^{1},\cdots, h_{m_{1}}^{1},\cdots,h_{1}^{|T|},h_{2}^{|T|},\cdots,h_{m_{|T|}}^{|T|}} \right\rbrack,
\end{equation}
in the first-level output of the model components, multiple different outputs are generated for the same task $t$, denoted as $h^t_{m_i}$. Here, $(1 \leq m_i \leq |m_t|)$ is used to identify the $m_i$-th output in the sequence of trained model components corresponding to the specific task $t$.
Our primary focus is on achieving knowledge fusion within the same task. Inspired by the MoE model proposed in~\cite{Shazeer2017}, we use the following expression to describe the fusion process:
\begin{equation}
  H = \left\lbrack {G_{i}(x) \cdot h^{*i}} \right\rbrack_{1 \leq i \leq {|T|}},
\end{equation}
here, ${G_{i}(x)}$ represents the task gating network for task $i$, with its output dimension matching the number of trained models for task $i$, denoted as $|m_{i}|$. Each output value of the task gating network ranges between 0 and 1, and the sum of all output values is always $1$. The symbol $h^{*i}$ represents the set of outputs from all model components for task $i$.
To obtain the final fused output $h^i$ for task $i$, we use each output of $G_{i}(x)$ as a weight and perform a weighted sum with the corresponding outputs in $h^{*i}$.
This method achieves knowledge fusion within the task. By performing this operation for each task, we obtain a set of outputs that have undergone intra-task knowledge fusion, denoted as $H = \left\lbrack {h^{1}, h^{2}, \cdots, h^{|T|}} \right\rbrack$.

After achieving effective integration of knowledge within the same task, we turn our attention to the issue of knowledge fusion between different tasks.
Although integrating all tasks at once may seem like the most straightforward approach, this method is prone to causing negative transfer of knowledge between tasks~\cite{Standley2020}. Such negative transfer can adversely affect the performance of each task, counteracting our original intent.
To mitigate the impact of negative transfer between tasks, we introduce the concept of temporal dependency between tasks~\cite{Xi2021}. This means that the output of one task may positively influence other tasks, although this influence is not always bidirectional. In a specific set of tasks, relying on manual methods to determine which tasks have such a promoting relationship and which task influences another is often unreliable. Therefore, we propose a Multi-Task Merge (MTM) method based on the self-attention mechanism.
In this method, we first use a fusion gating network to adaptively select two task outputs with a promoting relationship from the set $H$. This selection process can be described as follows:
\begin{equation}
  \left\lbrack {p_{i}^{x},p_{i}^{y}} \right\rbrack_{i \in {|T|}} = F_{i}(x) \odot H,
\end{equation}
here, $F_{i}(x)$ represents the fusion gating network for task $i$. It is a $|H|$-dimensional vector, with each dimension's output corresponding to a task in set $H$, and these output values range from 0 to 1.
For each task $i$, we equip it with a dedicated fusion gating network. By applying the operation $\odot$, we can filter out the maximum output value from $F_{i}(x)$, excluding the output of task $i$ itself. The task corresponding to this maximum value is considered the most closely related task to task $i$.
Next, we combine task $h^i$ with this selected related task, denoted as $\left\lbrack {p_{i}^{x}, p_{i}^{y}} \right\rbrack$.

After identifying these two related tasks, we perform a knowledge fusion operation specifically for them. The output calculation process of the MTM method is described as follows:
\begin{equation} 
  z^{i} = MTM\left( p_{i}^{x},p_{i}^{y} \right),
\end{equation}
using the MTM method, we can achieve adaptive fusion of knowledge related to specific tasks. The detailed fusion process is as follows:
\begin{equation} \label{5}
  z_{t} = {\sum\limits_{w \in (p_{i}^{x},p_{i}^{y})}{c \cdot V(w)}}, 
\end{equation}
where $c$ is the weight, which can be expressed as follows:
\begin{equation} \label{6}
  \begin{aligned}
  c &= SoftMax\left( \overline{c} \right), \\
  \overline{c} &= \frac{\left\langle {Q(w),K(w)} \right\rangle}{\sqrt{d}}.
  \end{aligned}
\end{equation}

In the above formula, $<\cdot,\cdot >$ represents a dot product, while $Q(\cdot )$, $K(\cdot )$, and $V(\cdot )$ are neural networks that project the input vectors onto a new vector representation. These functions, $Q(\cdot )$, $K(\cdot )$, and $V(\cdot )$, can be customized. In this paper, they are constructed using a simple fully connected network with an activation function. This approach bears resemblance to the self-attention mechanism~\cite{Vaswani2017}, where $Q(\cdot )$, $K(\cdot )$, and $V(\cdot )$ learn the Query, Key, and Value, respectively, from the input. The similarity between Query and Key is calculated using Equation~\ref{6}, where $d$ serves as a normalization factor. Then, the \emph{SoftMax} function is employed to transform these similarities into a probability distribution. Finally, the Value is weighted according to Equation~\ref{5}, a technique that has been previously validated in the existing studies~\cite{Xi2021a}.

\subsection{Efficient Multi-Task Modeling} \label{sec5}
Based on the deconstruction of trained models methods and AKF module, we are able to efficiently construct multi-task models. The specific process is demonstrated in Algorithm~\ref{alg2}.
\begin{algorithm}[htbp]
\SetAlgoLined %显示end
\caption{EMM}%算法名字
\label{alg2}
\KwIn{input data: $x$,
Model component set: $\left. M_{C}\leftarrow\left\lbrack {{M_{C}}_{1},{M_{C}}_{2},\cdots,{M_{C}}_{|m_c|}} \right\rbrack \right.$}%输入参数
\KwOut{$Y$}%输出
% some description\; %\;用于换行
\For{$m_l = {M_{C}}_{1},{M_{C}}_{2},\cdots,{M_{C}}_{|m_c|}$ \tcp*{Model components classified by levels}}{
\eIf{$m_l = {M_{C}}_{1}$ \tcp*{Hierarchical fusion}}{$Z = {KF}_{{M_{C}}_{1}}(x)$ }{$Z = {KF}_{m_l}(Z)$}
}
$Y = TOWER(Z)$ \tcp*{Obtaining the output set for all tasks}
\end{algorithm}

The algorithm receives input data $x$ and the set of model components $M_{C}$. The model components are categorized by hierarchical levels, with components at the same level sharing the same input and output. As shown in Algorithm~\ref{alg1}, each trained model can be deconstructed into $|m_c|$ hierarchical components.
In this algorithm, $KF_{m_l}$ represents the overall structure depicted in Figure~\ref{fig2} (where $m_l \in M_{C}$). When $l$ takes different values in $m_l$, it indicates that the model component part in Figure~\ref{fig2} consists of components from different hierarchical levels.
The input vector $x$ is first processed by the initial AKF module, generating an output $Z$. Then, $Z$ is used as the input for the second AKF module, and this process continues sequentially until the final AKF module outputs a $|T|$-dimensional vector $Z$. Subsequently, each dimension of $Z$ is fed into the corresponding task tower layer, resulting in the final output for each task.

\section{Experiments}
% \subsection{Dataset Description}
% \begin{itemize}[leftmargin=*]
% \item \textbf{Census-Income}: The dataset~\cite{2000} encompasses census income information from the United States, with the aim of predicting whether an individual's income exceeds \$50,000. It comprises 41 columns, including 7 dense features, 33 sparse features, and 1 label column to indicate income status. Following the precedent set by previous researchers~\cite{Tang2020}, we have established income prediction as the primary task, while also incorporating marital status prediction as an auxiliary task.
% \item \textbf{Ali-CCP (Alibaba Click and Conversion Prediction)}: The dataset~\cite{Ma2018a} originates from the logs of the mobile Taobao application. Specifically, it comprises 23 sparse features and 8 dense features, accompanied by two crucial labels: "click" and "purchase".
% \item \textbf{AliExpress (AliExpress Searching System Dataset)}: Utilizing data~\cite{Li2020} from the U.S. region of the Alibaba's AliExpress platform, and the data features can be divided into 16 sparse features and 63 dense features. It encompasses three crucial labels: "exposure," "click," and "conversion", providing a significant basis for analyzing user behavior and market trends.
% \end{itemize}
\subsection{Dataset Description}
\begin{itemize}[leftmargin=*]
\item \textbf{Census-Income}~\cite{2000}: U.S. census income data, predicting if income exceeds \$50,000. 41 columns: 7 dense, 33 sparse features, 1 label. Primary task: income prediction. Auxiliary task: marital status prediction~\cite{Tang2020}.
\item \textbf{Ali-CCP}~\cite{Ma2018a}: From mobile Taobao logs. 23 sparse, 8 dense features. Labels: "click", "purchase".
\item \textbf{AliExpress}~\cite{Li2020}: From Alibaba's AliExpress (U.S.). 16 sparse, 63 dense features. Labels: "exposure", "click", "conversion". Aids user behavior and market trend analysis.
\end{itemize}
% The statistical data of these datasets are shown in Table~\ref{tab1}.
% \begin{table}[htbp]
%   \centering
%   \caption{Summary Statistics of the Dataset.}
%   \label{tab1}
%   \resizebox{\columnwidth}{!}{
%   \begin{threeparttable}
%   \begin{tabular}{@{}c|cccccc@{}}
%   \toprule
%   \textbf{Dataset} & \textbf{Task} & \multicolumn{1}{l}{\textbf{Train}} & \multicolumn{1}{l}{\textbf{Validation}} & \textbf{Test} & \multicolumn{1}{l}{\textbf{DF$^\ast$}} & \multicolumn{1}{l}{\textbf{SF$^\ast$}} \\ \midrule
%   Census-Income & 1 & 199K & 49K & 49K & 7  & 32 \\
%   Ali-CCP       & 2 & 42M  & 21M & 21M & 8  & 23 \\
%   AliExpress    & 3 & 19M  & -   & 7M  & 63 & 16 \\ \bottomrule
%   \end{tabular}
%   \begin{tablenotes}
%   \scriptsize \item[$\ast$] DF: Dense Features; SF: Sparse Features
%   \end{tablenotes}            %这行要添加
%   \end{threeparttable} 
%   }
%   \end{table}
We employ datasets of varying sizes and with differing proportions of data features to validate the effectiveness of our proposed methodology.

\subsection{Evaluation Protocol}

In the experimental phase, to comprehensively evaluate the performance of the proposed EMM method and baseline models, we followed the evaluation criteria selected by Ma et al.~\cite{Ma2018a, Ma2018} and used AUC as the primary evaluation metric. AUC, or the area under the Receiver Operating Characteristic (ROC) curve, has the unique advantage of being unaffected by decision thresholds, compared to other evaluation metrics such as accuracy, recall, and F1 score. The calculation formula for AUC can be expressed as:
\begin{equation}
\begin{split}
  AUC &= \frac{1}{N_{+}}{\sum\limits_{j = 1}^{N_{+}}\left( {r_{j} - j} \right)}/N_{-} \\
&= \frac{\sum_{j = 1}^{N_{+}}{r_{j} - N_{+}\left( N_{+} + 1 \right)/2}}{N_{+}N_{-}}
\end{split}
\end{equation}
% The calculation principle of AUC is based on the ranking of samples. Specifically, the model's predicted positive and negative samples are sorted in ascending order by their scores. For the $j$-th positive sample, if its rank is $r_j$, it means there are $r_j-1$ samples before it, among which $j-1$ are positive samples. Therefore, the number of negative samples ranked before the $j$-th positive sample (with scores lower than it) is $r_j-j$. In other words, for the $j$-th positive sample, the probability that its score is higher than a randomly selected negative sample is $(r_j-j)/N_{-}$, where $N_{-}$ represents the total number of negative samples. The larger the AUC value, the stronger the model's ability to distinguish between positive and negative samples.

% To ensure the accuracy and comprehensiveness of the evaluation results, this study calculates and reports the AUC values for each task across all datasets. For different datasets, we focus on two core tasks for performance evaluation: for the Census-Income dataset, we primarily evaluate the income and marital status tasks; for the Ali-CCP dataset, we focus on the prediction of clicks and purchases; and for the AliExpress dataset, we concentrate on the prediction accuracy of clicks and conversions.
AUC calculation ranks model's positive/negative samples by scores. For $j$-th positive sample ranked $r_j$, it has $r_j-j$ negatives before it. Its score's probability of exceeding a random negative is $(r_j-j)/N_{-}$, where $N_{-}$ is total negatives. Higher AUC indicates better model discrimination.

This study calculates AUC for each task across datasets. Evaluations focus on core tasks: Census-Income (income, marital status), Ali-CCP (clicks, purchases), AliExpress (click, conversion accuracy).

\subsection{Baseline Methods and Model Settings}
\begin{table*}[htbp]
\centering
\caption{The AUC Results Across Three Datasets}
\label{tab2}
\resizebox{\textwidth}{!}{%
\begin{threeparttable}
\begin{tabular}{@{}l|ccll|ccll|ccll@{}}
\toprule
\multicolumn{1}{c|}{\multirow{2}{*}{\textbf{Model}}} &
  \multicolumn{4}{c|}{\textbf{Census-Income}} &
  \multicolumn{4}{c|}{\textbf{Ali-CCP}} &
  \multicolumn{4}{c}{\textbf{AliExpress}} \\
\multicolumn{1}{c|}{} &
  T1-AUC &
  T2-AUC &
  \multicolumn{2}{c|}{Gain} &
  T1-AUC &
  T2-AUC &
  \multicolumn{2}{c|}{Gain} &
  T1-AUC &
  T2-AUC &
  \multicolumn{2}{c}{Gain} \\ \midrule
TM-1 &
  0.93009 &
  0.99230 &
  \multicolumn{1}{c}{-} &
  \multicolumn{1}{c|}{-} &
  0.60862 &
  0.62297 &
  \multicolumn{1}{c}{-} &
  \multicolumn{1}{c|}{-} &
  0.86524 &
  0.70310 &
  \multicolumn{1}{c}{-} &
  \multicolumn{1}{c}{-} \\
TM-2 &
  0.90739 &
  0.99356 &
  \multicolumn{1}{c}{-} &
  \multicolumn{1}{c|}{-} &
  0.61937 &
  0.62324 &
  \multicolumn{1}{c}{-} &
  \multicolumn{1}{c|}{-} &
  0.86958 &
  0.70880 &
  \multicolumn{1}{c}{-} &
  \multicolumn{1}{c}{-} \\ \midrule
PLE &
  0.92528 &
  0.99465 &
  +0.00654 &
  +0.00172 &
  0.57442 &
  0.57975 &
  -0.03957 &
  -0.04335 &
  {\ul 0.81261} &
  0.69853 &
  {\ul -0.05481} &
  -0.00741 \\
MMoE &
  0.94051 &
  0.99437 &
  +0.02176 &
  +0.00143 &
  {\ul 0.58419} &
  0.58951 &
  {\ul -0.02980} &
  -0.03360 &
  0.80499 &
  {\ul 0.69988} &
  -0.06243 &
  {\ul -0.00606} \\
AITM &
  \textbf{0.94527} &
  0.99436 &
  \textbf{+0.02653} &
  +0.00143 &
  0.57866 &
  0.58389 &
  -0.03533 &
  -0.03921 &
  0.78133 &
  0.69943 &
  -0.08608 &
  -0.00652 \\
AdaTT &
  {\ul 0.94523} &
  \textbf{0.99479} &
  \multicolumn{1}{c}{{\ul +0.02649}} &
  \multicolumn{1}{c|}{\textbf{+0.00186}} &
  0.58076 &
  {\ul 0.60080} &
  \multicolumn{1}{c}{-0.03324} &
  \multicolumn{1}{c|}{{\ul -0.02231}} &
  0.70414 &
  0.69111 &
  \multicolumn{1}{c}{-0.16327} &
  \multicolumn{1}{c}{-0.01484} \\
Ours &
  0.94108 &
  {\ul 0.99467} &
  +0.02234 &
  {\ul +0.00174} &
  \textbf{0.59916} &
  \textbf{0.61597} &
  \textbf{-0.01484} &
  \textbf{-0.00714} &
  \textbf{0.85080} &
  \textbf{0.70571} &
  \textbf{-0.01661} &
  \textbf{-0.00024} \\ \bottomrule
\end{tabular}%
\begin{tablenotes}
\footnotesize
\item In Census-Income dataset, T1: income, T2: marital status.
In Ali-CCP dataset, T1: purchase, T2: click.
In AliExpress dataset, T1: conversion, T2: click.
\end{tablenotes}            %这行要添加
\end{threeparttable}
}
\end{table*}

We compare our method with mainstream models:
\begin{itemize}[leftmargin=*]
\item \textbf{Trained Models}: Use MLPs (TM-1, TM-2) as single-task models. For Census-Income, use 8- and 16-layer MLPs for two tasks each (4 models total). For other datasets, use 128- and 256-layer MLPs consistently.
\item \textbf{MMoE}~\cite{Ma2018}: Uses expert networks. Decompose each model into 3 components (12 total for EMM framework). Configure 12 expert networks matching components. For Census-Income, set 8 hidden layers in experts; for others, 128.
\item \textbf{PLE}~\cite{Tang2020}: Based on expert networks, with task-specific and shared experts. Construct two-layer CGC network with 12 experts (4 task-specific, 2 shared per layer), similar to MMoE.
\item \textbf{AITM}~\cite{Xi2021}: Models temporal task relationships. Design shared bottom with same expert structure as PLE for comparability.
\item \textbf{AdaTT}~\cite{Li2023}: Deep fusion network with task-specific and shared units, using residual/gating for inter-task fusion. Configure 12 experts consistent with MMoE/PLE.
\end{itemize}

Experiments use a server with two RTX 4090 GPUs and PyTorch. Settings adjusted per dataset. All use Adam optimizer. For Census-Income: batch size 1024, learning rate 1e-3, weight decay 1e-6, 100 epochs. For Ali-CCP/AliExpress: batch size 32,768, learning rate 1e-3, weight decay 1e-5, 10 epochs.

\subsection{Performance Comparison}
In this subsection, we will present the AUC scores of all models on the test set and further analyze their gains. The gains mentioned here refer to the results obtained by comparing with the average AUC of TM-1 and TM-2.
Bold text indicates the best results among methods other than the trained models, while underlined text indicates the second-best results.
For the Census-Income dataset, we primarily examined two tasks: income and marital status. The pertinent results are detailed in Table~\ref{tab2}. Through an in-depth analysis of these data, we have arrived at the following insightful observations:
\begin{itemize}[leftmargin=*]
\item All multi-task models outperform the average performance of single-task models on both tasks, indicating that integrating multi-task information within the model can provide additional performance improvements. Project models based on expert bottom layers, such as PLE and MMoE, are designed to enhance model performance by precisely controlling information sharing between different tasks. Unfortunately, these models do not perform well on the Census-Income dataset.
\item The AITM method constructs knowledge transfer strategies based on temporal relationships between tasks, achieving excellent results on Task 1. This demonstrates that the AITM method can effectively utilize and transfer knowledge to improve model performance. However, it is worth noting that its knowledge transfer is unidirectional and requires manual setup. While this enhances the performance of the primary task, it may lead to relatively weaker performance in auxiliary tasks.
The AdaTT method achieves the best performance on Task 2, benefiting from its ability to adaptively integrate task knowledge.
\item Our proposed method ranks third on Task 1 and second on Task 2. Although the overall performance is not outstanding and slightly lags behind AITM, which focuses on temporal relationships, and AdaTT, which focuses on knowledge fusion, the gap is minimal, with only a 0.44\% and 0.01\% decrease, respectively.
\end{itemize}

The experimental results for the Ali-CCP and AliExpress datasets are similar and are therefore discussed together as follows:
\begin{itemize}[leftmargin=*]
\item Compared to single-task models, all multi-task models exhibit a performance decline. This phenomenon may be attributed to factors such as the increased scale of the datasets, the imbalanced distribution of sample categories, and potential negative transfer effects between tasks.
\item Our proposed EMM method achieves the best performance across all tasks on both datasets, strongly validating the effectiveness of the method. The EMM method facilitates knowledge transfer between similar and different tasks, effectively mitigating adverse factors in multi-task models.
\item The AITM and AdaTT methods only achieve one second-best result across the four tasks in the two datasets. This indicates that as the dataset scale increases, relying solely on temporal relationship strategies or adaptive task knowledge fusion strategies is insufficient to achieve optimal results.
\item The MMoE and PLE methods, based on expert bottom layers, achieve three second-best results across the four tasks in the two datasets. This demonstrates that the mixture of experts approach can exhibit relatively superior performance in the context of increasing data volume and complexity.
\end{itemize}

\subsection{Ablation Study}

In this subsection, we conduct an ablation study on the MTM method within the AKF module and the trained models. The specific steps are as follows:
First, we use a baseline method, which maintains the original model structure without including the trained models and the MTM method. We train this baseline model from scratch and record the AUC values for its tasks.
Second, based on the baseline model, we introduce the MTM method but do not include the trained models. We retrain the model with the MTM method integrated and similarly record the AUC values.
Next, we return to the baseline model and this time, we include the trained models without the MTM method. We retrain this model to obtain its AUC values.
Finally, we conduct a detailed comparative analysis of the above three methods with our proposed method. Our proposed method integrates both the trained models and the MTM method. The results of the ablation study are shown in Table~\ref{tab3}.
\begin{table*}[h]
  \centering
  \caption{Experimental Results of Ablation Study}
  \label{tab3}
  \begin{threeparttable}
  \resizebox{\textwidth}{!}{%
  \begin{tabular}{@{}c|cccc|cccc|cccc@{}}
  \toprule
  \multirow{2}{*}{\textbf{Model}} &
    \multicolumn{4}{c|}{\textbf{Census-Income}} &
    \multicolumn{4}{c|}{\textbf{Ali-CCP}} &
    \multicolumn{4}{c}{\textbf{AliExpress}} \\
    &
    T1-AUC &
    T2-AUC &
    \multicolumn{2}{c|}{Gain} &
    T1-AUC &
    T2-AUC &
    \multicolumn{2}{c|}{Gain} &
    T1-AUC &
    T2-AUC &
    \multicolumn{2}{c}{Gain} \\ \midrule
  baseline &
    0.94088 &
    0.99383 &
    - &
    - &
    0.52897 &
    0.61216 &
    - &
    - &
    0.76183 &
    0.70159 &
    - &
    - \\ \midrule
  baseline+MTM &
    {\ul 0.93215} &
    {\ul 0.99376} &
    -0.00873 &
    -0.00007 &
    {\ul 0.54757} &
    {\ul 0.61483} &
    +0.01860 &
    +0.00267 &
    {\ul 0.84987} &
    0.69863 &
    +0.08804 &
    -0.00296 \\
  baseline+p &
    0.92647 &
    0.99352 &
    -0.01441 &
    -0.00031 &
    0.53139 &
    0.58600 &
    +0.00242 &
    -0.02616 &
    0.71285 &
    {\ul 0.69965} &
    -0.04898 &
    -0.00194 \\
  Ours &
    \textbf{0.94108} &
    \textbf{0.99467} &
    \textbf{+0.0002} &
    \textbf{+0.00084} &
    \textbf{0.59916} &
    \textbf{0.61597} &
    \textbf{+0.07019} &
    \textbf{+0.00381} &
    \textbf{0.85080} &
    \textbf{0.70571} &
    \textbf{+0.08897} &
    \textbf{+0.00412} \\ \bottomrule
\end{tabular}%
}
\begin{tablenotes}
\footnotesize 
\item p: trained model.
\end{tablenotes}            %这行要添加
\end{threeparttable} 
\end{table*}

% Gain represents the difference between the AUC value of the model and the baseline.
% Based on these study results, we draw the following meaningful conclusions:

% \begin{itemize}[leftmargin=*]
% \item Intuitively, utilizing trained model components for multi-task modeling should enhance model performance or at least maintain its original level. However, experimental results show that simply combining these trained components does not significantly impact model performance and may even lead to performance degradation. The reason for this phenomenon is that once a model loses its original structure, the dimensionality of its output information may become incompatible with the new model, necessitating the introduction of new modules to interpret and transform this information.
% \item Although introducing the MTM method can slightly improve model efficiency, its enhancement over the baseline model is not significant. This is mainly because the core of the MTM method lies in knowledge transfer, and when the correlation between tasks is weak, its efficiency improvement is naturally limited.
% \item We observe that the EMM method proposed in this study demonstrates the best efficiency. This finding reveals a complementary effect between the MTM method in the AKF module and the trained model components; only their combined action can significantly improve the performance of multi-task models. It is worth noting that relying on either of these methods alone cannot effectively enhance the efficiency of multi-task models.
% \end{itemize}
Gain is the difference between a model's AUC and the baseline, and our study reveals that combining trained model components for multi-task modeling does not significantly enhance performance and may even degrade it due to output dimensionality mismatches. While the MTM method slightly improves efficiency, its impact is limited when task correlation is weak. Notably, the EMM method demonstrates the best efficiency, highlighting a complementary effect between MTM in the AKF module and trained model components, as only their combined use significantly improves multi-task model performance.

\subsection{Adaptability Analysis}

\begin{figure*}[htbp]  % htbp表示图片的排版位置  
  \centering  % 使图片居中  
  \begin{subfigure}[b]{0.32\textwidth}  % b表示bottom对齐，0.3\textwidth表示子图的宽度  
  \centering  
  \includegraphics[width=\textwidth]{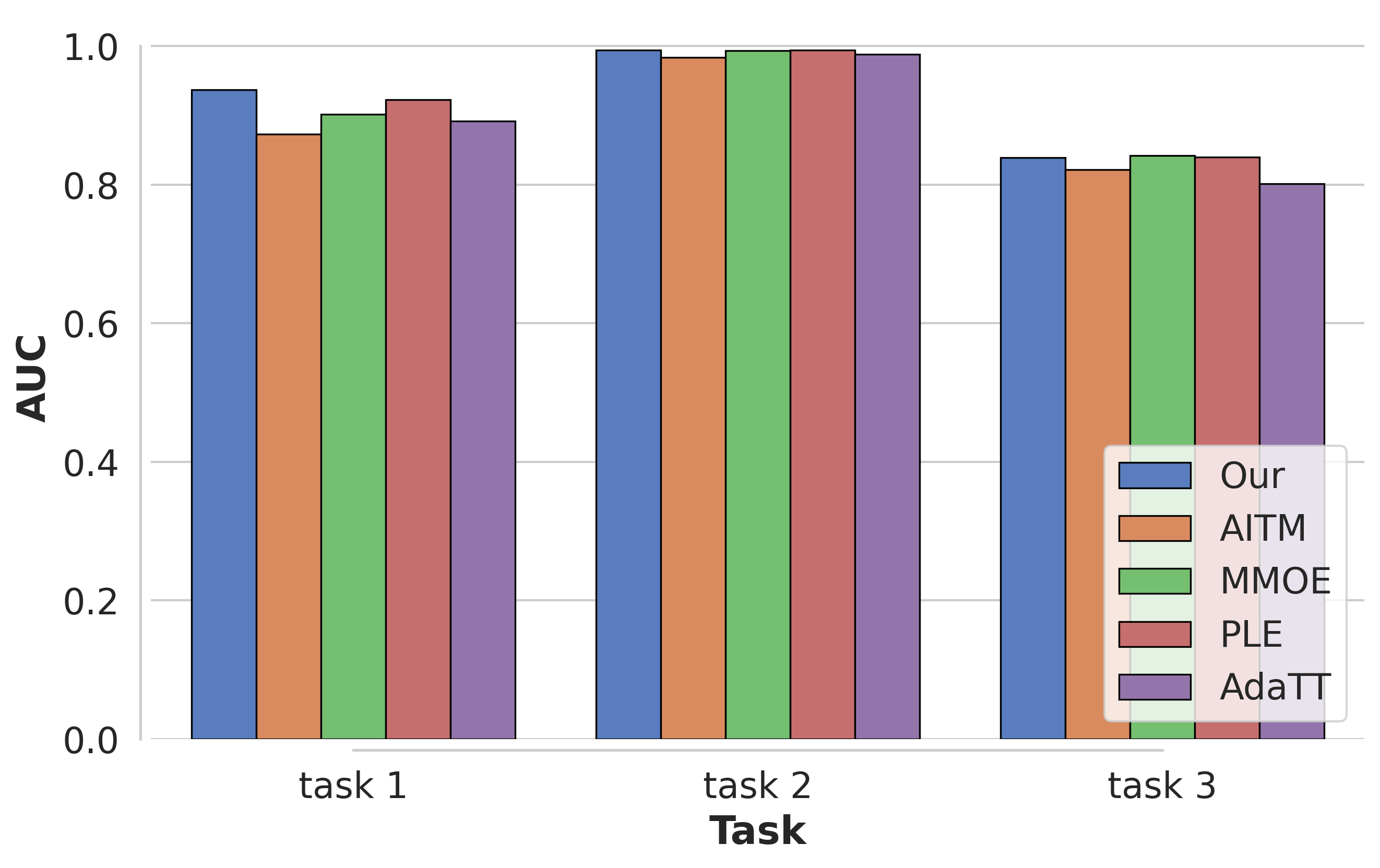}  % 插入图片，width=\textwidth表示图片宽度与子图宽度一致  
  \caption{3 tasks performed on Census-Income}  % 子图标题  
  \end{subfigure}
  \hfill  % 填充水平空间，使子图之间有一定的间距  
  \begin{subfigure}[b]{0.32\textwidth}
  \centering
  \includegraphics[width=\textwidth]{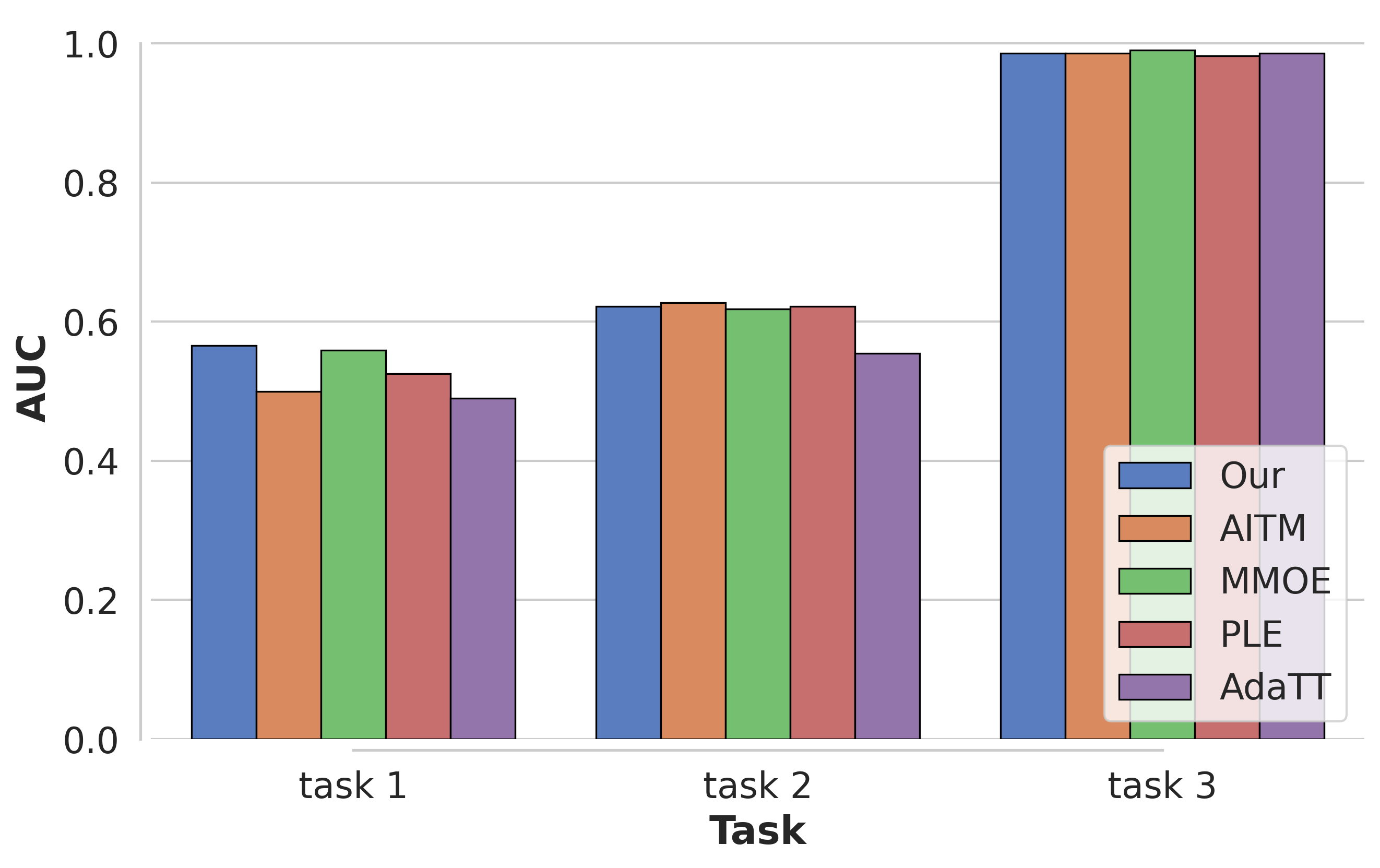}
  \caption{3 tasks performed on Ali-CCP}
  \end{subfigure}
  \hfill
  \begin{subfigure}[b]{0.32\textwidth}
  \centering
  \includegraphics[width=\textwidth]{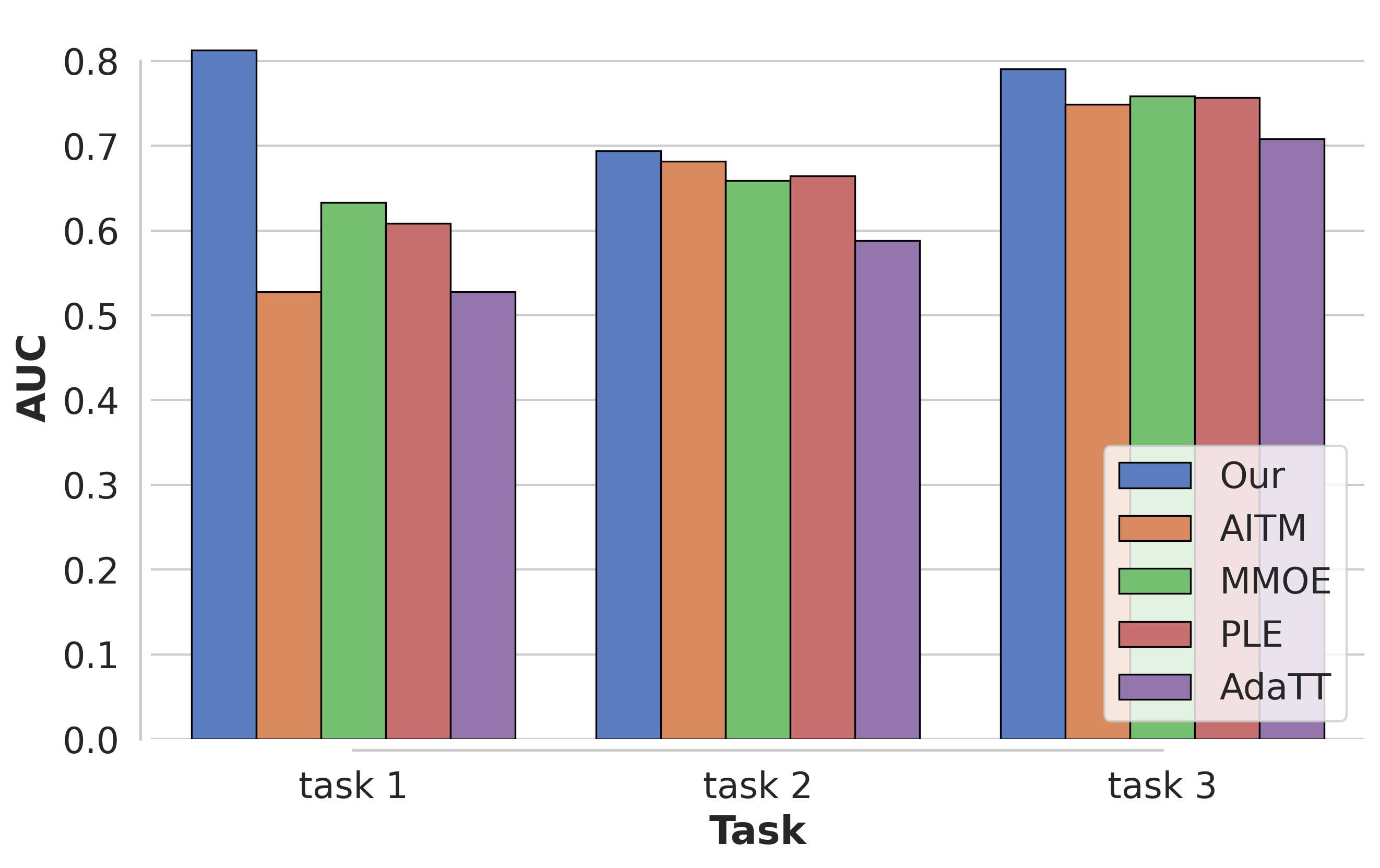}  
  \caption{3 tasks performed on AliExpress}
  \end{subfigure}  
  
  \vskip\baselineskip  % 垂直间距，使两行子图之间有一定的间距  
  
  \begin{subfigure}[b]{0.32\textwidth}  
  \centering  
  \includegraphics[width=\textwidth]{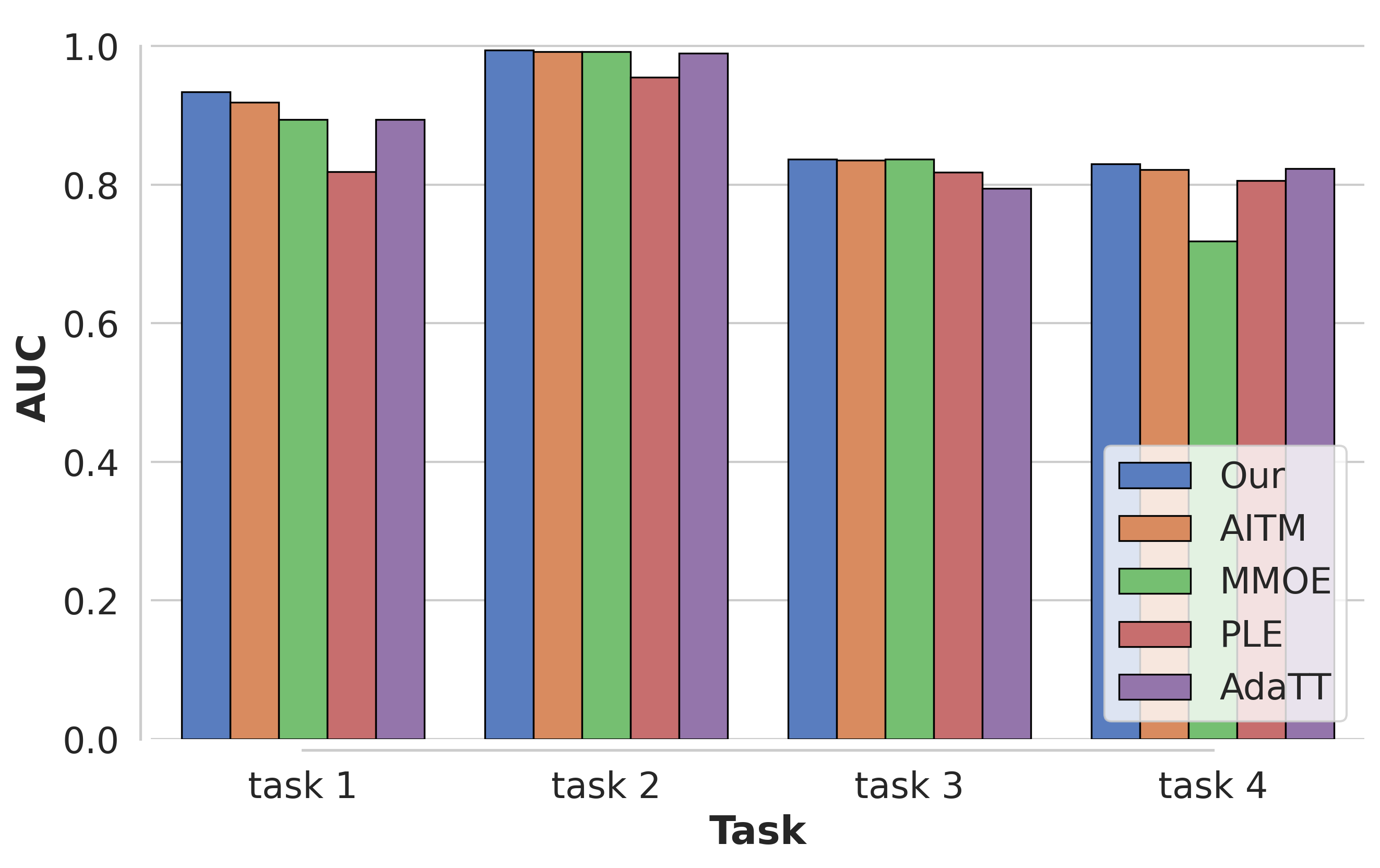}  
  \caption{4 tasks performed on Census-Income}
  \end{subfigure}  
  \hfill  
  \begin{subfigure}[b]{0.32\textwidth}  
  \centering  
  \includegraphics[width=\textwidth]{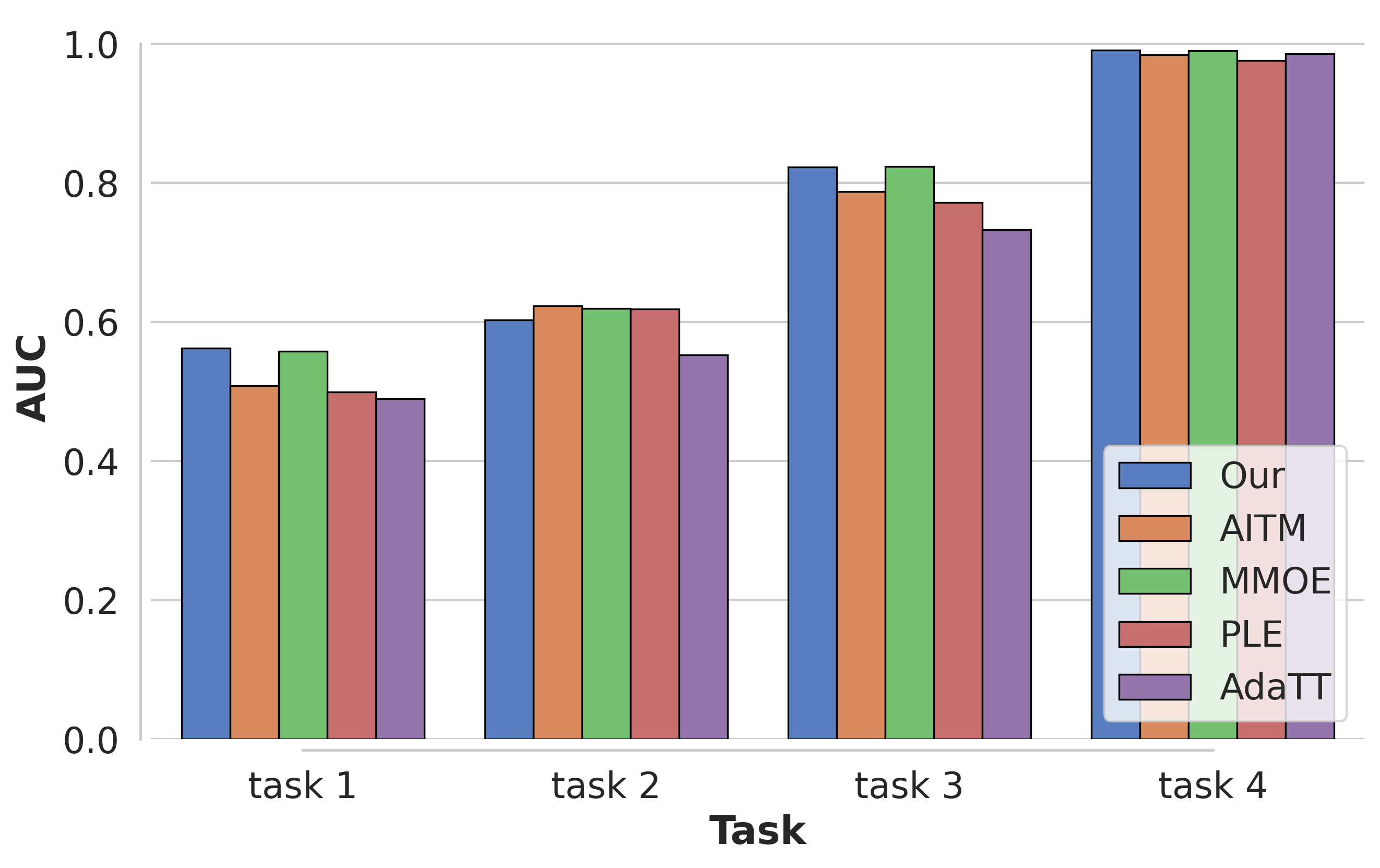}  
  \caption{4 tasks performed on Ali-CCP}  
  \end{subfigure}  
  \hfill  
  \begin{subfigure}[b]{0.32\textwidth}  
  \centering  
  \includegraphics[width=\textwidth]{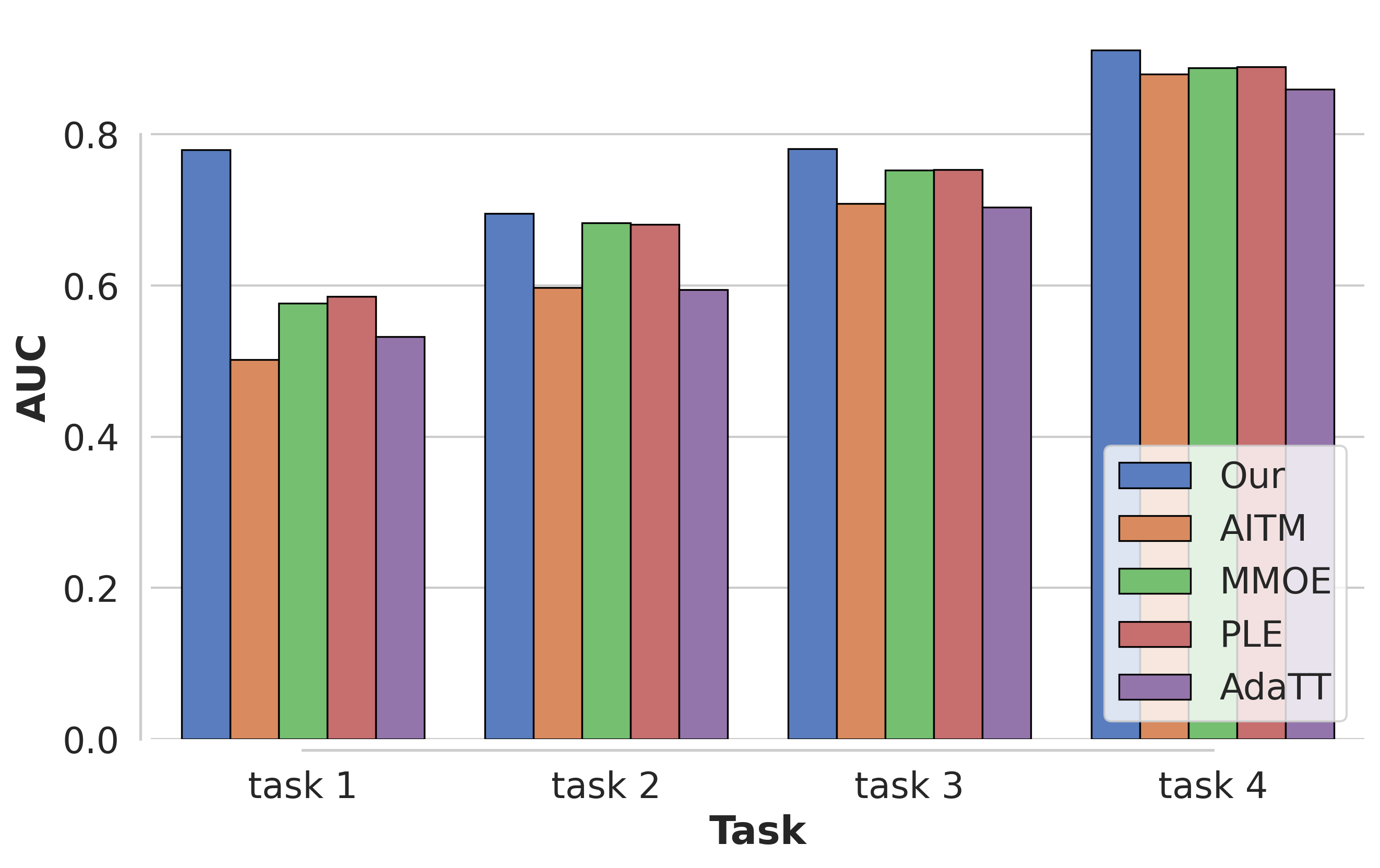}  
  \caption{4 tasks performed on AliExpress}  
  \end{subfigure}  
  \caption{To test the adaptability of the EMM method for multi-task processing, we increased the number of tasks. In the experimental results charts, the X-axis represents the different tasks, while the Y-axis shows the AUC evaluation values.} \label{fig3} % 主图标题  
  \end{figure*}

Adaptability analysis is a key aspect of evaluating the EMM method's response to changing modeling conditions. This experiment focuses on the core element of the modeling process—the number of tasks—by adjusting this variable to verify the adaptability of the EMM method. The experimental results, as shown in Figure~\ref{fig3}, visually present the performance of the EMM method under different numbers of tasks.

The experimental results show that as the number of tasks increases, our proposed method consistently demonstrates superior performance. Specifically, across the six task scenarios covered by the three datasets, the EMM method only slightly underperforms in Task 2 of the Ali-CPP dataset, while achieving optimal or leading performance in all other tasks. Notably, in the AliExpress dataset, the performance advantage of the EMM method becomes increasingly significant with the addition of more tasks, clearly surpassing other methods.

\section{Conclusion}
This paper proposes an efficient multi-task modeling method—EMM, which aims to achieve an automated modeling solution by integrating trained models. Multi-task modeling often involves complex task relationship analysis and cumbersome modeling processes. To simplify this process, we innovatively utilize trained models and achieve multi-task modeling through the automated fusion of these models. Specifically, the method first decomposes trained models into multiple components, then employs an AKF module based on the self-attention mechanism to effectively integrate these components hierarchically. Finally, by stacking the AKF modules in layers, a new multi-task model is constructed. This method not only retains the performance advantages of trained models but also achieves efficient and automated multi-task modeling. To validate the effectiveness of the EMM method, we conducted comprehensive experimental studies on three typical datasets. The results show that the EMM method significantly outperforms traditional multi-task modeling methods across various task configurations and datasets. Overall, this method eliminates the need for manually parsing task relationships and modeling steps, opening a new pathway for efficiently constructing multi-task models with its superior performance and automated modeling capabilities.

\balance
\bibliographystyle{IEEEtran}
\normalem
\bibliography{reference}

\end{document}